\documentclass[runningheads]{llncs}
\usepackage[T1]{fontenc}
\usepackage{multicol,multirow}
\usepackage{url}
\usepackage{booktabs}
\usepackage[colorlinks, linkcolor=blue, anchorcolor=blue, citecolor=blue]{hyperref}
\usepackage{graphicx,verbatim}
\usepackage{amssymb}
\usepackage{makecell} 
\usepackage{floatflt} 
\usepackage{wrapfig}
\usepackage{caption}
\usepackage{cleveref}
\usepackage[misc]{ifsym} 

\begin{document}
\title{Empowering Medical Multi-Agents with Clinical Consultation Flow for Dynamic Diagnosis}
%

\author{Sihan Wang$^1$, Suiyang Jiang$^2$, Yibo Gao$^1$, Boming Wang$^1$, Shangqi Gao$^3$, Xiahai Zhuang$^1$$^{(\textrm{\Letter})}$}  
\authorrunning{Sihan Wang et al.}
\institute{$^1$ The School of Data Science, Fudan University,\\
$^2$Institute of Science and Technology for Brain-Inspired Intelligence, Fudan University, \\
$^3$the Department of Oncology, University of Cambridge\\
    \email{zxh@fudan.edu.com}}

\maketitle              
\begin{abstract}
Traditional AI-based healthcare systems often rely on single-modal data, limiting diagnostic accuracy due to incomplete information. However, recent advancements in foundation models show promising potential for enhancing diagnosis combining multi-modal information. While these models excel in static tasks, they struggle with dynamic diagnosis, failing to manage multi-turn interactions and often making premature diagnostic decisions due to insufficient persistence in information collection.
To address this, we propose a multi-agent framework inspired by consultation flow and reinforcement learning (RL) to simulate the entire consultation process, integrating multiple clinical information for effective diagnosis. Our approach incorporates a hierarchical action set, structured from clinic consultation flow and medical textbook, to effectively guide the decision-making process. This strategy improves agent interactions, enabling them to adapt and optimize actions based on the dynamic state. We evaluated our framework on a public dynamic diagnosis benchmark. The proposed framework evidentially improves the baseline methods and achieves state-of-the-art performance compared to existing foundation model-based methods.

\end{abstract}
\section{Introduction}\label{Section: introduction}
\begin{figure}[t]
    \centering
\includegraphics[width=1\linewidth]{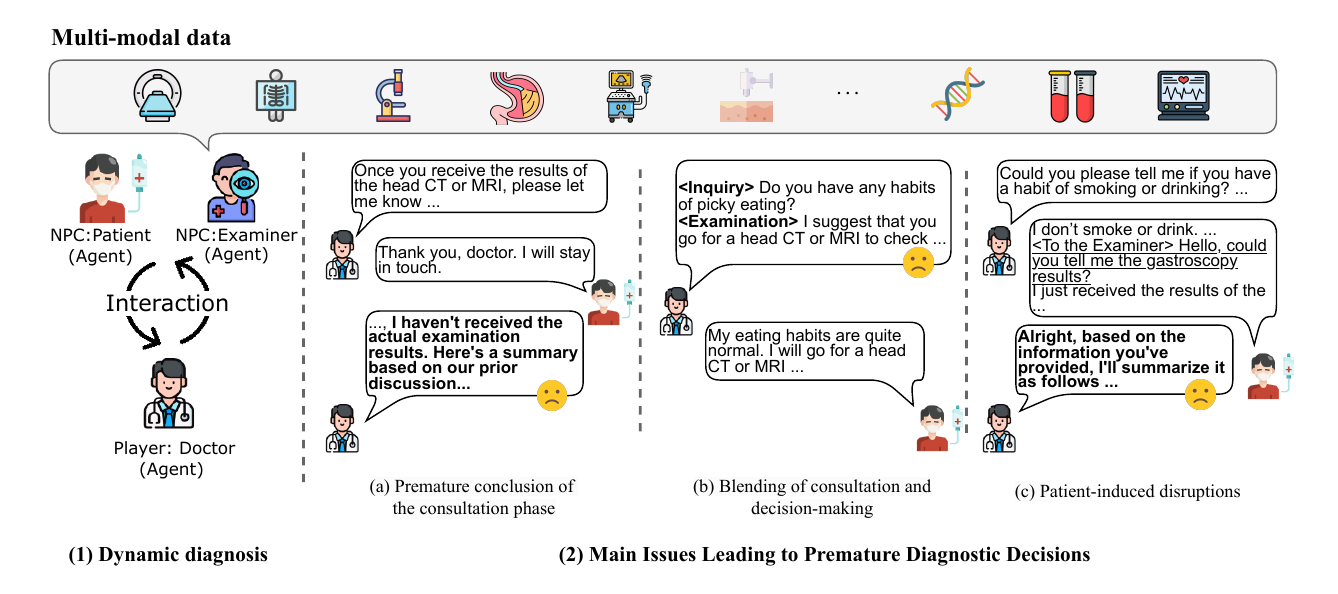}
    \caption{(1) Illustration of multi-agent dynamic diagnosis framework. (2) Main issues resulting in recent \emph{premature closure} challenge in the interaction. }
    \label{fig:motivation}
\end{figure}

AI-based healthcare has traditionally utilized single-modal data, such as medical images or laboratory results, which limits decision-making due to incomplete information~\cite{multi-modal}. Advances in foundation models now allow for the integration of multiple modalities~\cite{multi-modal}, demonstrating great potential in medical applications~\cite{med_llava}. Despite promising results in static benchmarks~\cite{zuo2025medxpertqa}, these models still struggle to emulate the dynamic nature of real-world clinical consultations~\cite{ai_hospital}.
Therefore, in this work, we focus on a more complex challenge, namely, dynamic diagnosis. It requires designing a multi-agent framework to simulate clinic consultation processes \cite{ai_hospital}, as illustrated in Fig. \ref{fig:motivation} (1). 
The framework generally includes several agents, such as the \emph{doctor}, \emph{patient}, and \emph{examiner}. Specifically, the \emph{doctor} agent engages in multi-turn interactions with the \emph{patient} to gather information, recommend appropriate tests, and make a final diagnosis. Meanwhile, the examiner handles multi-modal data, including medical images, ECGs, and laboratory test results, to extract and relay objective information to other agents. Inspired by advances in multi-modal foundation model, we integrate all information into the text space, as language is a natural medium for enhancing interpretability and human-AI interaction \cite{FAMA}. In this work, we leverage well-developed foundation models, with a primary focus on optimizing the multi-agent interaction flow.

While foundation models have shown promising performance in multi-modal static diagnosis, they struggle with multi-turn interactions, leading to limited diagnostic performance \cite{ai_hospital}. A key challenge is the tendency of foundation models to make \emph{premature closure} \cite{premature_closure}, where they formulate and confirm hypotheses too quickly, conducting diagnoses based on incomplete information, which can lead to inaccuracies. This issue is also a significant concern in clinical practice, affecting human doctors \cite{premature_closure}. We identify three main factors that contribute to suboptimal diagnoses, as illustrated in Fig. \ref{fig:motivation}: (a) the premature conclusion of the consultation phase; (b) the blending of consultation and diagnostic decision-making; and (c) patient-induced disruptions. These issues lead to fewer interaction turns, with most sessions having only 4-5 turns (as shown in Fig. \ref{fig:exp} (a)), showing a limited persistence in data collection, which is a major factor in assessing a human doctor's consultation performance \cite{premature_closure}. Consequently, this leads to premature decisions and reduced diagnostic accuracy.

To tackle \emph{premature closure}, we use reinforcement learning (RL) to regularize the solution space, helping the foundation model focus on reasoning and select the best action based on the current state \cite{CoELA}. Although multi-agent RL frameworks have been explored in other fields \cite{RL_survey,RL_work,CoELA}, research in dynamic diagnosis is still limited. We propose an RL-driven multi-agent framework that integrates clinical consultation knowledge with the reasoning capabilities of foundation model. Specifically, we design a hierarchical action set derived from medical textbooks to guide the solution space and ensure that the foundation model focuses on selecting the most appropriate actions.

Building on the success of RL methods, we introduce an RL-driven, multi-agent framework for dynamic diagnosis. Our contributions are as follows: 
\begin{itemize}
    \item We introduce a novel medical multi-agent framework that incorporates clinical consultation flow, where the \emph{doctor} agent selects the optimal action from a formulated set based on the current consultation state. 
    \item We design a hierarchical action set derived from clinical consultation processes and medical textbooks, ensuring a structured decision-making pipeline for the \emph{doctor} agent. 
    \item We evaluate our framework with several foundation models on a public dynamic diagnosis benchmark, significantly improving baseline performance and achieving state-of-the-art results.
\end{itemize}

\section{Methodology}
\begin{figure}[t]
    \centering
\includegraphics[width=\linewidth]{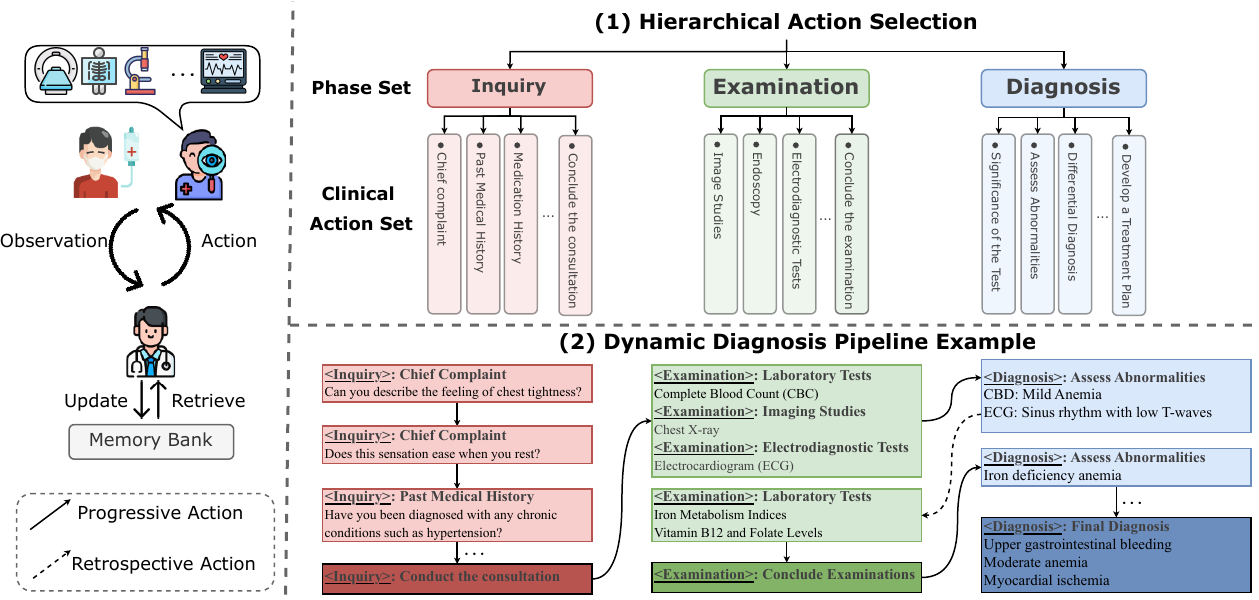}
    \caption{(1) Illustration of the proposed hierarchical action set derived from consultation flow. (2) An example of the Dynamic diagnosis pipeline.}
    \label{fig:pipeline}
\end{figure}




\subsection{Dynamic Diagnosis Formulation}
Dynamic diagnosis could be formulated as a partially observable Markov decision game (POMG)~\cite{POMDP}, involving three agents: \emph{doctor}, \emph{patient}, and \emph{examiner}. The interaction between the agents in dynamic diagnosis can be denoted as $(n,S, A, O, T, G, R)$. Specifically, $n=3$ refers to the number of agents and $S$ denotes a set of the diagnosis process states. $A$ is an action set for the \emph{doctor} agent, formulated from the clinical consultation process. $O$ is the observation set for the \emph{doctor} agent, which consists of the subjective observation from the \emph{patient} agent (\textit{i.e.}, symptoms or complaints from patients) and the objective observation from the \emph{examiner} agent, (\textit{i.e.}, the result of imaging examinations or lab tests). 
$T(s,a,s') = p(s'|s,a)$ is the joint transition model that defines the probability that achieving the new diagnosis state $s'$ after taking the joint action $a$ in $s$; $G=\{g_1, ...,g_k\}$ defines the task with several sub-goals for the \emph{doctor} agent to finish, and the ultimate goal is to achieve the precise diagnosis. The sub-goal will be detailed in the next section. Moreover, $R$ is the reward function.
In this formulation, the \emph{doctor} agent uses the available observations to select the best action for an accurate diagnosis. The partially observable nature of the game reflects that the agent might not have access to all information at once, and thus must make decisions based on partial or uncertain data, much like in real-world clinical settings.

\subsection{Incorporating Clinical Consultation Flow}
To mitigate the issue of premature closure in prior frameworks~\cite{ai_hospital}, we enhance the multi-agent framework by incorporating a real-world clinical consultation flow. The consultation process is divided into three major phases: inquiry, examination, and diagnosis. Each phase is associated with a corresponding sub-goal $g_k$ for the \emph{doctor} agent, ensuring a more structured approach to dynamic diagnosis:
\begin{itemize}
    \item \textbf{Inquiry Phase}: The \emph{doctor} gathers subjective symptoms or complaints from the \emph{patient} to form an initial diagnosis.
    \item \textbf{Examination Phase}: The \emph{doctor} selects appropriate tests to gather objective data, confirming or challenging the initial diagnosis.
    \item \textbf{Diagnosis Phase}: The \emph{doctor} analyzes the examination results, refine their diagnosis, and make a final decision on the patient's condition.
\end{itemize}

The \emph{doctor} agent has distinct objectives in each phase, which in turn leads to different action spaces. As such, it is necessary to constrain the action space for each phase. Medical textbooks clearly define the key components of the consultation process, such as the present medical history and the history of the present illness. To this end, we leverage GPT to summarize the action points from medical textbooks \cite{Textbook} and formulate a clinical action set for each phase.

\subsection{Hierarchical Action Selection by the AI Agent}

Given a state $s$, the \emph{doctor} agent needs to select the optimal actions based on the current state. To achieve it effectively, the \emph{doctor} agent requires a robust reasoning module. While developing such a module from scratch would consume large efforts, we leverage the advanced reasoning capabilities of foundation model. As shown in Fig. \ref{fig:pipeline}, the \emph{doctor} agent retrieves patient-related information from the memory bank, compiles an action list based on the current state, and incorporates prior medical knowledge stored in the foundation model to guide decision-making.

The action selection process of the AI agent follows a hierarchical structure. First, the \emph{doctor} agent selects an appropriate consultation phase from a predefined set based on the current state. Subsequently, the corresponding clinical action set is identified. The chosen actions are then translated into detailed test descriptions that align with the current context. For instance, a typical action might be "\texttt{<Inquiry>: Chief Complaint. Do you feel headache?}". A demo of this process is shown in Fig. \ref{fig:pipeline} (2).
This action selection strategy not only efficiently regularizes the solution space but also facilitates stage transitions when necessary. In Fig. \ref{fig:pipeline} (2), two types of actions are presented: \textit{progressive action} and \textit{retrospective action}. The former moves the process forward through the diagnostic steps, while the latter allows the system to backtrack to a prior phase, and add additional required tests. Together, these actions enable a more thorough and adaptive consultation process.

\section{Experiments}
\subsection{Dataset and Evaluation Metrics} \label{section: data}
We utilized the public benchmark, namely MVME \cite{ai_hospital} for evaluation. It consists of 506 complete medical cases, including case reports and paired multi-modal information (such as medical images, ECG, and laboratory tests).
We utilized two kinds of evaluation strategies to estimate the consultation performance, following the MVME benchmark \cite{ai_hospital}, \textit{i.e.}, Score evaluation on agent-generated consultation report, and Match-based evaluation of diagnostic results. 
\textbf{Score evaluation} \cite{ai_hospital}: We introduce an additional agent to score the generated report (0-100) across five aspects: Symptoms, Medical Examinations, Diagnostic Results, Diagnostic Rationales, and Treatment Plan. The agent has access to the complete medical record, i.e., the ground truth (GT). \emph{Note that, in this case, we use GPT-4o as the evaluator, which results in slightly different baseline results compared to those in the MVME, where GPT-4 was used. }
\textbf{Match-based evaluation} \cite{ai_hospital}: We evaluate diagnostic results using entity-overlap-based automated metrics. Disease entities are extracted from both agent-generated diagnostics and ground-truth medical records, then mapped to standardized ICD-10 disease entities~\cite{ICD-10}. Entity overlap is computed to assess the final diagnoses. The evaluation metrics include set-level precision, recall, and F1 score.

\subsection{\emph{Doctor} Agent Behavior Analysis in Dynamic Diagnosis} \label{experiment:Agent behavior}
In this section, we present the comparison results from two perspectives. We implemented our proposed framework with two recent foundation models (Qwen-Max and Deepseek-V3), denoted as Ours (Qwen-Max) and Ours (Deepseek), respectively. The results are compared with the baseline performance of five well-known foundation models: GPT-3.5, Wenxin-4.0, GPT-4, Qwen-Max, and Deepseek-V3. 

Table~\ref{table:exp} [Score evaluation] presents the comparison results in five aspects. Our framework significantly improved the performance of both baselines. Specifically, Ours (Qwen-Max) achieved the highest scores across all metrics, outperforming the baseline (Qwen-Max) by at least 10 points in all five aspects. Ours (Deepseek-V3) also achieved significant improvements, achieving the second-best performance. 

\begin{table}[!tb]
\caption{Comparison of Model Performance with and without our RL strategy. GPT-4+GT$^*$ refers to the upper bound, where the ground truth for symptoms and medical examinations is provided. R and P refer to Recall and Precision. }
\centering
\resizebox{\linewidth}{!}{
\begin{tabular}{l|ccccc|ccc}
\toprule
& \multicolumn{5}{c|}{Score evaluation} & \multicolumn{3}{l}{Match-based evaluation }\\
\hline
Model  & Symptoms & \makecell{Medical\\Examination} & \makecell{Diagnostic\\Result} & \makecell{Diagnostic\\Rationales} & \makecell{Treatment\\Plan} & R & P & F1 \\ \hline
GPT-3.5 &  58.50$_{\pm 1.71}$  &  32.74$_{\pm 2.70}$  &  26.09$_{\pm 2.70}$  &  27.60$_{\pm 2.70}$ &  15.68$_{\pm 1.98}$ &  19.19 &  37.39 &  25.37 \\ 
Wenxin-4.0 &  59.88$_{\pm 1.52}$  &  29.64$_{\pm 2.64}$  &  25.43$_{\pm 2.50}$  &  26.61$_{\pm 2.70}$ &  19.63$_{\pm 2.24}$ &  22.03 &  31.44 &  25.91 \\ 
GPT-4 &  60.41$_{\pm 1.38}$  &  29.45$_{\pm 2.37}$  &  31.56$_{\pm 2.77}$  &  32.74$_{\pm 2.70}$ &  24.70$_{\pm 2.24}$ &  21.64 &  50.26 &  30.26 \\ 
Deepseek-V3 &  65.22$_{\pm 1.52}$  &  36.43$_{\pm 2.64}$  &  36.56$_{\pm 2.83}$  &  39.59$_{\pm 2.77}$ &  29.78$_{\pm 2.37}$  &  24.78 &  50.21 &  33.18 \\ 
Qwen-Max &  64.89$_{\pm 1.78}$ &  35.24$_{\pm 2.64}$ &  32.87$_{\pm 2.83}$ &  36.43$_{\pm 2.83}$ &  27.27$_{\pm 2.37}$ &  22.42 &  43.38 &  29.56 \\ 
\hline
Ours (Deepseek-V3) &  74.97$_{\pm 1.45}$  &  61.33$_{\pm 1.78}$  &  46.71$_{\pm 2.77}$  &  50.86$_{\pm 2.57}$  &  42.16$_{\pm 2.31}$ &  31.68 &  \textbf{50.92} &  39.06 \\ 
Ours (Qwen-Max) &  \textbf{76.55$_{\pm 1.52}$}  &  \textbf{63.31$_{\pm 1.98}$}  &  \textbf{49.80$_{\pm 2.57}$}  &  \textbf{56.00$_{\pm 2.44}$}  &  \textbf{45.92$_{\pm 2.44}$} &  \textbf{33.41} &  50.61 &  \textbf{40.25} \\ 
\hline
GPT-4+GT$^*$ &  100.0*  &  100.0*  &  58.89$_{\pm 1.65}$  &  66.60$_{\pm 1.32}$  &  53.16$_{\pm 1.91}$ &  38.90 &  58.97 &  46.88 \\ 
\bottomrule
\end{tabular}
}
\label{table:exp}
\end{table}

Table~\ref{table:exp} [Match-based evaluation] presents the entity-overlap-based evaluation of diagnostic results. Our framework significantly enhanced diagnostic performance. Specifically, Ours (Qwen-Max) achieves the highest performance, with Recall and F1 scores of 33.41, and 40.25, respectively, and Ours (Deepseek-V3) achived the best performance with Precision of 50.92. This demonstrates that through the proposed strategy, the \emph{doctor} agent can gather more information, leading to better diagnostic outcomes.

\begin{table}[!tb]
\caption{Ablation study of applying our RL strategy to individual phases. Here the \emph{baseline} refers to Qwen-Max, and the \emph{Ours-Stage} represents applying our RL strategy to the specific stage (inquiry, examination and diagnosis).}
\centering
\resizebox{0.80\linewidth}{!}{
\begin{tabular}{l|ccccc}
\toprule
Models    & Symptoms & \makecell{Medical\\ Examinations} & \makecell{Diagnostic\\ Results} & \makecell{Diagnostic\\ Rationales} & \makecell{Treatment\\ Plan} \\
\midrule
Baseline & 64.89$_{\pm 1.78}$ & 35.24$_{\pm 2.64}$ & 32.87$_{\pm 2.83}$ & 36.43$_{\pm 2.83}$ & 27.27$_{\pm 2.37}$ \\
\midrule
Ours-inquiry   &    69.17$_{\pm 1.58}$      &      \underline{45.85}$_{\pm 2.57}$                &         36.56$_{\pm 2.83}$           &       41.83$_{\pm 2.70}$                &     32.35$_{\pm 2.37}$           \\
Ours-examination &     \underline{69.70}$_{\pm 1.52}$     &     42.95$_{\pm 2.77}$                 &   \underline{38.21}$_{\pm 2.77}$                 &        \underline{43.54}$_{\pm 2.77}$               &   \underline{33.73}$_{\pm 2.50}$             \\
Ours-diagnosis  &     68.71$_{\pm 1.65}$     &     38.87$_{\pm 2.64}$                 &           36.30$_{\pm 2.70}$         &      40.84$_{\pm 2.83}$                 &     31.95$_{\pm 2.50}$           \\
\midrule
Ours-All     &     \textbf{76.55$_{\pm 1.52}$}     &      \textbf{63.31$_{\pm 1.98}$}                &  \textbf{49.80$_{\pm 2.57}$}                  &        \textbf{56.00$_{\pm 2.44}$}               &    \textbf{45.92$_{\pm 2.44}$}               \\
\bottomrule
\end{tabular}
}
\label{table:ablation}
\end{table}

\subsection{Impact of Individual Phase}
We assess the impact of RL strategies on each consultation stage (inquiry, examination, and diagnosis), implementing them with Qwen-Max as the baseline agent, and each model is denoted as Ours-stage (e.g., Ours-inquiry).


Table~\ref{table:ablation} presents the performance results of the score evaluation. Each model achieved improvements compared to the baseline. Ours-inquiry achieved the greatest improvement in Medical Examinations, highlighting the importance of effective inquiry in selecting appropriate tests. Ours-examination demonstrated the most significant improvement across four out of five metrics but ranks second in Medical Examinations, due to the impact of incomplete inquiry on the examination phase. Nonetheless, thorough examination results provide critical data, enhancing the understanding of the patient’s condition and leading to better diagnostic outcomes. In contrast, Ours-diagnosis achieved only a slight improvement, emphasizing that while diagnosis is important, effective inquiry and examination are crucial for accurate, informed decision-making. Without sufficient data from these earlier stages, diagnostic reasoning becomes less effective.

\begin{figure}[t]
    \centering
    \includegraphics[width=0.85\linewidth]{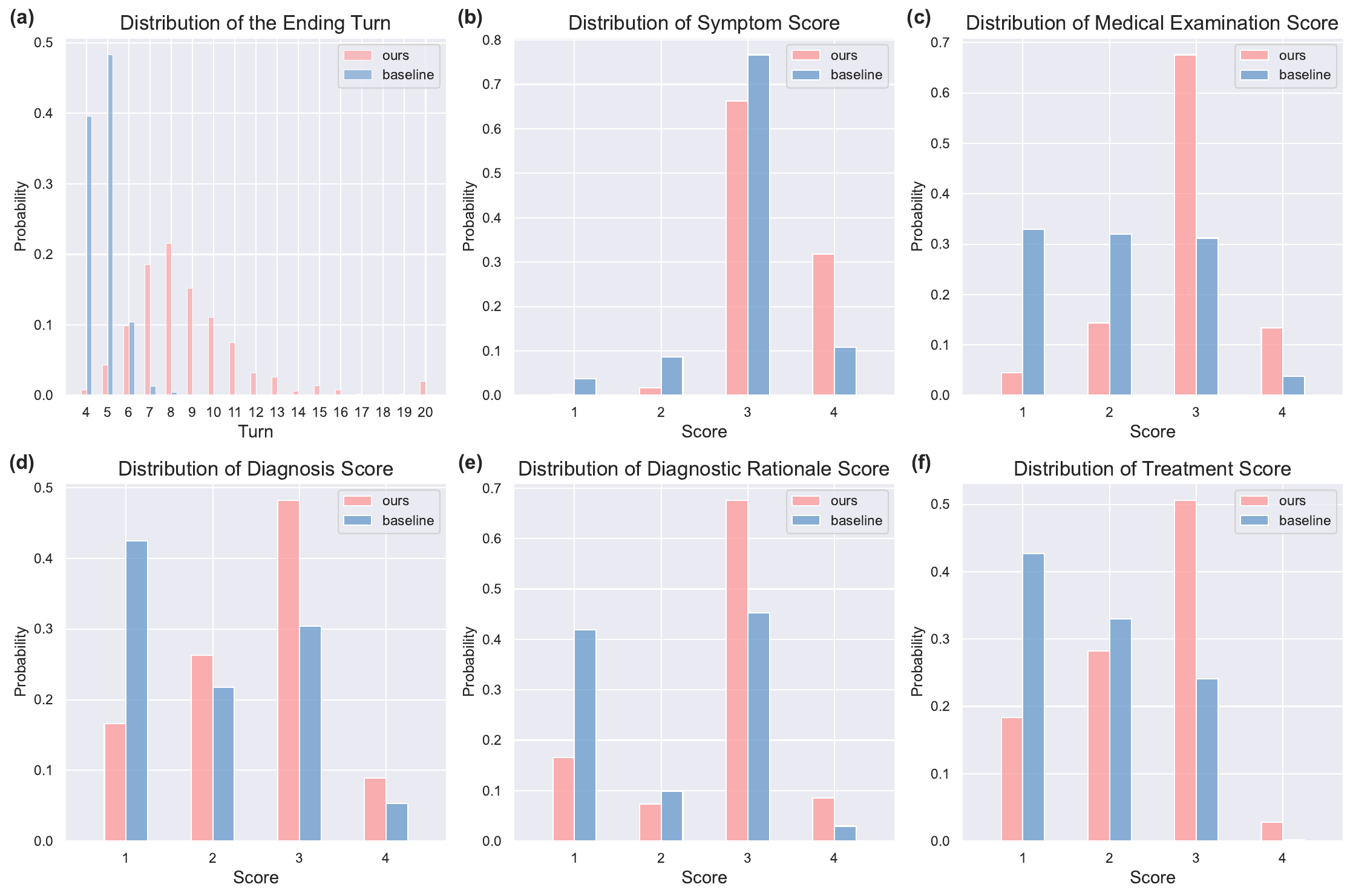}
    \caption{Distribution of (a) dynamic diagnosis interaction turns and (b)-(f) evaluation scores in the 5 aspects.}
    \label{fig:exp}
\end{figure}

\subsection{Further Analysis and Case Study}
In this section, we analyze the results of our framework and present three case studies to demonstrate its performance in addressing the issues outlined in Section \ref{Section: introduction}, using Ours (Qwen-Max) as an example (denoted as Ours in this section).

Figure \ref{fig:exp} illustrates the effectiveness of our framework compared to the baseline, in the persistence in data collection and metric scores. 
Fig. \ref{fig:exp} (a) shows the frequency distribution of interaction turns. The baseline results are concentrated around 4-5 turns, indicating limited information gathering by the \emph{doctor} agent and minimal adaptation to different patients. In contrast, Ours achieved a higher mode of 8 turns with greater variance, reflecting its ability to adapt the interaction duration based on patient needs. This flexibility enables Ours to gather more comprehensive information, leading to better diagnostic performance. Fig. \ref{fig:exp} (b-f) compares Ours and the baseline across the five metrics from Table~\ref{table:exp}, showing that Ours improved performance by reducing lower scores and increasing higher ones, resulting in an overall boost across all metrics.

\begin{figure}[t]
    \centering
    \includegraphics[width=\linewidth]{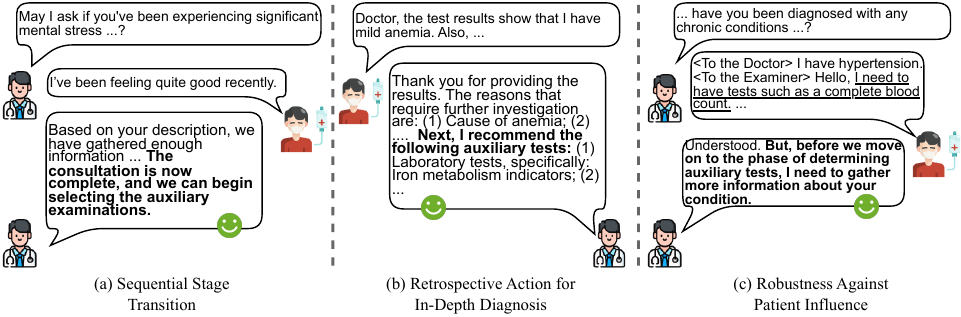}
    \caption{Case Study. An illustration of how the proposed framework mitigates the challenge of premature closure.}
    \label{fig:case study}
\end{figure}

Fig. \ref{fig:case study} presents three cases illustrating how the proposed framework mitigates the
challenge of \emph{premature closure}. Fig. \ref{fig:case study} (a) refers to the sequential stage transition, that is, the \emph{doctor} follows the consultation flow, ensuring that it moves to the next phase only after sufficient information has been gathered. Fig. \ref{fig:case study} (b) illustrates the retrospective action for in-depth diagnosis, where the doctor gathers additional information, resulting in a more thorough diagnosis and reducing premature conclusions. In Fig. \ref{fig:case study} (c), it demonstrate the robustness against patient influence, namely, the \emph{doctor} remains unaffected by \emph{patient} influence, maintaining the integrity of the diagnostic process without disruptions.

In summary, our framework prevents premature closure and allows the \emph{doctor} to gather more information, improving diagnostic accuracy over the baseline.

\section{Conclusion}
In this work, we proposed a multi-agent framework powered by foundation model and reinforcement learning to address challenges in dynamic diagnosis. By formulating a hierarchical action set, our framework enables a more comprehensive diagnostic process. The incorporation of consultation knowledge and reinforcement learning ensures that the \emph{doctor} can prevent premature closure, and improving diagnostic accuracy. Our results demonstrate significant improvements over baseline methods, highlighting the effectiveness of multi-modal integration and dynamic, interactive decision-making in real-world clinical settings. 

Future work will extend this framework by introducing more agents, particularly multi-modal agents, allowing for better diagnostic outcomes through agent interactions, collaboration, and even competition among the agents.

\clearpage
\bibliographystyle{splncs04}

\end{document}